\documentclass{article}

\usepackage[main, nonatbib, final]{neurips_2025}
\usepackage[utf8]{inputenc} % allow utf-8 input
\usepackage[T1]{fontenc}    % use 8-bit T1 fonts
\usepackage{url}            % simple URL typesetting
\usepackage{booktabs}       % professional-quality tables
\usepackage{amsfonts}       % blackboard math symbols
\usepackage{nicefrac}       % compact symbols for 1/2, etc.
\usepackage{microtype}      % microtypography
\usepackage[table]{xcolor}

\usepackage{amsmath}
\usepackage{amssymb}
\usepackage{mathtools}
\usepackage{amsthm}

\usepackage[pagebackref,breaklinks,colorlinks]{hyperref}
\usepackage{multirow}
\usepackage[ruled,linesnumbered]{algorithm2e}
\usepackage{wrapfig}

\newtheorem{theorem}{Theorem}[section]
\newtheorem{proposition}[theorem]{Proposition}

\newtheorem{definition}[theorem]{Definition}

\DeclareMathOperator*{\argmax}{arg\,max}
\newcommand{\olsi}[1]{\,\overline{\!{#1}}}
\title{Siegel Neural Networks}

\author{%
  Xuan Son Nguyen \quad Aymeric Histace \quad Nistor Grozavu \\%\thanks{Use footnote for providing further information
  ETIS, UMR 8051, CY Cergy Paris University, ENSEA, CNRS\\
  \texttt{\{xuan-son.nguyen,aymeric.histace\}@ensea.fr} \\
  \texttt{nistor.grozavu@cyu.fr} \\
}

\begin{document}

\maketitle

\begin{abstract}
Riemannian symmetric spaces (RSS) such as hyperbolic spaces and symmetric positive definite (SPD) manifolds have become popular spaces 
for representation learning. In this paper, we propose a novel approach for building discriminative neural networks on Siegel spaces, 
a family of RSS that is largely unexplored in machine learning tasks. 
For classification applications, one focus of recent works is the construction of multiclass logistic regression (MLR) 
and fully-connected (FC) layers for hyperbolic and SPD neural networks. 
Here we show how to build such layers for Siegel neural networks. 
Our approach relies on the quotient structure of those spaces and the notation of vector-valued distance on RSS. 
We demonstrate the relevance of our approach on two applications, i.e., radar clutter classification and node classification. 
Our results successfully demonstrate state-of-the-art performance across all datasets. 
\end{abstract}

%%%%%%%%% BODY TEXT ----------------------------------------------------------------------------------------------------------
\section{Introduction}
\label{sec:intro}

Deep neural networks are generally built upon the assumption that the data or features 
at hand exhibit Euclidean latent structure. 
Unfortunately, this assumption does not hold in many applications~\cite{arsigny:inria-00070423,NEURIPS2018_dbab2adc} 
where the data or features lie on a multidimensional curved surface which is locally Euclidean.  
For such applications, those models often produce unsatisfactory results because their building blocks based on Euclidean geometry  
break the geometric stability principle that plays a crucial role in geometric deep learning architectures~\cite{bronstein2021geometricdeeplearninggrids}. 
To deal with this issue, many Riemannian neural networks have been developed for 
solving a wide variety of 
machine learning problems~\cite{ZChengICLR25,NEURIPS2018_dbab2adc,HuangGool17,HuangAAAI18,katsman2023riemannian,shimizu2021hyperbolic}. 
In this paper, we restrict our attention to discriminative neural networks with manifold-valued output. 
 
Early works focus either on hyperbolic spaces~\cite{NEURIPS2018_dbab2adc,shimizu2021hyperbolic} 
or on matrix manifolds~\cite{HuangGool17,Huang17DLLieGroup,HuangAAAI18}. 
In an attempt to develop a unified framework for a more general setting, 
the authors of~\cite{NguyenNeurIPS22,NguyenGyroMatMans23,NguyenICLR24} leverage  
the gyro-structure of certain Riemannian manifolds. 
However, in the general case, their methods cannot provide an explicit form for the point-to-hyperplane distance 
which is at the heart of their proposed network building blocks~\cite{shimizu2021hyperbolic} 
since the distance must be derived with respect to a specific Riemannian metric. 
The work in~\cite{NguyenICLR25} alleviates this issue by deriving 
a closed form for the point-to-hyperplane distance associated with $G$-invariant Riemannian metrics on RSS. 
Although this work is applicable to Siegel spaces, 
the construction of the MLR and FC layers~\cite{NguyenICLR25} from the derived distance is heavily based on the maximal abelian 
subspaces of those spaces. This can affect the ability of the resulting networks to learn rich representations 
and complex decision boundaries.  

In this paper, we propose a novel approach for building neural networks on Siegel spaces. 
Those are among the most versatile RSS~\cite{FedericoSymspaces21} and have many attractive theoretical properties. 
The two well-established models of Siegel spaces, i.e., the Siegel upper space and the Siegel disk~\cite{SiegelSymGeometry} 
generalize the complex Poincaré upper plane and the complex Poincaré disk~\cite{helgason1979differential} to spaces of 
symmetric complex matrices~\cite{NielsenSiegelKleinDisk20}. 
SPD manifolds associated with affine-invariant Riemannian metrics~\cite{pennec:tel-00633163} are also 
special cases of Siegel spaces associated with Siegel metrics~\cite{SiegelSymGeometry}. 
Despite the potential of Siegel spaces in capturing rich geometrical structures, 
they are much less studied in the context of deep learning compared to other RSS. Although few recent works~\cite{FedericoSymspaces21,taha2023normedspacesgraphembedding} use Siegel spaces as representation spaces,  
they only focus on learning and visualizing embeddings in natural language processing and graph tasks. 
Therefore, an effective framework for building discriminative Siegel neural networks is still missing.  
In summary, our contributions are the following:
\begin{itemize}
\item We propose a novel formulation of MLR layers for Siegel neural networks based on a gyrovector approach 
which has proven effective in building hyperbolic and SPD neural networks~\cite{NEURIPS2018_dbab2adc,NguyenNeurIPS22,shimizu2021hyperbolic}. 
This formulation is then extended to the product space setting. 
\item We show that the notion of vector-valued distance~\cite{kapovich2017anosovsubgroupsdynamicalgeometric} 
which captures the complete $G$-congruence invariant of pairs of points on RSS enables another formulation of 
MLR layers for Siegel neural networks in a natural way. This formulation leads to more compact MLR layers than those obtained 
by the first formulation. 
\item We introduce two variants of FC layers for Siegel neural networks. 
\item We build the first discriminative Siegel neural networks and evaluate them on the radar clutter classification and node classification tasks. 
\end{itemize}

%-----------------------------------------------------------------------------------------------------------------------------
\section{Mathematical Background}
\label{sec:mathematical_background}

%--------------------------------------------------------------------------------------------------
\subsection{Siegel Spaces}
\label{sec:siegel_spaces_symspaces}

The Siegel upper space $\mathbb{SH}_m$ is defined as %the space of symmetric complex square
%matrices %of size $m \times m$ 
%which have positive-definite imaginary part
\begin{equation*}
\mathbb{SH}_m = \{ x = u + iv: u \in \operatorname{Sym}_m, v \in \operatorname{Sym}_m^+ \},
\end{equation*}
where $\operatorname{Sym}_m$ and $\operatorname{Sym}_m^+$ denote the space of $m \times m$ real symmetric matrices 
and that of $m \times m$ SPD matrices, respectively.

Another model for Siegel spaces is the Siegel disk defined by
\begin{align*}
\begin{split}
\mathbb{SD}_m &= \left \{ w \in \operatorname{Sym}_{m,\mathbb{C}}: I_m - ww^H \in \mathbb{H}^+_m \right \}, 
\end{split}
\end{align*}
where $I_m$ is the $m \times m$ identity matrix, 
$\operatorname{Sym}_{m,\mathbb{C}}$ and $\mathbb{H}^+_m$ denote the space of $m \times m$ complex symmetric matrices and 
that of $m \times m$ Hermitian positive definite (HPD) matrices, respectively, 
and $w^H$ is the conjugate transpose of $w$.

One can convert a point $x \in \mathbb{SH}_m$ to $\mathbb{SD}_m$ using the matrix Cayley transformation defined as
\begin{equation*}
\varphi(x) = (x - iI_m)(x + iI_m)^{-1}.
\end{equation*}

The inverse matrix Cayley transformation that converts a point $w \in \mathbb{SD}_m$ to $\mathbb{SH}_m$ is given by
\begin{equation*}
\varphi^{(-1)}(w) = i(I_m + w)(I_m - w)^{-1}.
\end{equation*}

In the following, we shall focus on the Siegel upper space model. 

\paragraph{Quotient Structure}

Denote by
\begin{align*}
\begin{split}
%\begin{equation*}
\operatorname{Sp}_{2m} = \left \{ \begin{bmatrix} a & b \\ c & d \end{bmatrix}: ab^T=ba^T,  cd^T=dc^T, ad^T - bc^T = I_{m} \right \}
%\end{equation*}
\end{split}
\end{align*}
the real symplectic group. 
This group acts transitively on $\mathbb{SH}_m$ by the action
$s[x] = (ax+b)(cx+d)^{-1}$, 
where $s = \begin{bmatrix} a & b \\ c & d \end{bmatrix} \in \operatorname{Sp}_{2m}$ and $x \in \mathbb{SH}_m$. 
The stabilizer group of $x = iI_m \in \mathbb{SH}_m$ is the subgroup of symplectic orthogonal matrices $\operatorname{SpO}_{2m}$ defined as:
\begin{equation*}
\operatorname{SpO}_{2m} = \left \{ \begin{bmatrix} a & b \\ -b & a \end{bmatrix}: a^Ta + b^Tb = I_m, a^Tb \in \operatorname{Sym}_m \right \} = \operatorname{Sp}_{2m} \cap \operatorname{O}_{2m},
\end{equation*}
where $\operatorname{O}_{2m}$ is the group of orthogonal matrices. 
We thus have the identification $\mathbb{SH}_m \cong \operatorname{Sp}_{2m} / \operatorname{SpO}_{2m}$. 
The element in $\operatorname{Sp}_{2m}$ that transforms $iI_m$ to $x = u + iv \in \mathbb{SH}_m$ via the group action
is given by the map $\phi(\cdot)$ in the following identification:
\begin{align*}
\begin{split}
\psi: \mathbb{SH}_m & \rightarrow \operatorname{Sp}_{2m}/\operatorname{SpO}_{2m} \\ x & \mapsto \begin{bmatrix} v^{\frac{1}{2}} & uv^{-\frac{1}{2}} \\ \mathbf{0} & v^{-\frac{1}{2}} \end{bmatrix} \operatorname{SpO}_{2m} = \phi(x) \operatorname{SpO}_{2m}.
\end{split}
\end{align*}

\paragraph{Riemannian Metric}

The Riemannian metric (also referred to as Symplectic metric) for the Siegel upper space model is given~\cite{SiegelSymGeometry} by
\begin{equation}\label{eq:siegel_line_element}
ds^2_x = 2 \operatorname{Tr}(v^{-1}dxv^{-1}d\olsi{x}), 
\end{equation}
where $x = u + iv \in \mathbb{SH}_m$, and $\operatorname{Tr}(\cdot)$ is the matrix trace (see Appendix~3.2 for further discussions). 
The associated Riemannian distance $d_{\mathbb{SH}}(x,y)$ between two points $x, y \in \mathbb{SH}_m$ is given by
\begin{equation*}
d_{\mathbb{SH}}(x,y) = \sqrt{\sum_{j=1}^m \log^2 \left ( \frac{1 + r_j^{\frac{1}{2}}}{1 - r_j^{\frac{1}{2}}} \right )},
\end{equation*}
where $r_j,j=1,\ldots,m$ are the eigenvalues of the cross-ratio $R(x,y)$ defined as 
\begin{equation*}
R(x,y) = (x-y)(x-\olsi{y})^{-1}(\olsi{x}-\olsi{y})(\olsi{x}-y)^{-1},
\end{equation*}
where $\olsi{x}$ denotes the complex conjugate of $x$. 

%--------------------------------------------------------------------------------------------------
\subsection{Riemannian Symmetric Spaces} 
\label{sec:symmetric_space_theory}

This section briefly reviews key concepts from the theory of RSS for our work. 
We refer the interested reader to Appendix~3.1 for further discussions. 

A symmetric space is a connected Riemannian manifold $X$ with a geodesic-reversing isometry at each point. 
In other words, for each point $x \in X$ there is an isometry $\sigma_x$ of $X$ such that $\sigma_x(x) = x$ 
and the differential of $\sigma_x$ at $x$ is multiplication by $-1$~\cite{bridson2011metric}. 
Siegel spaces belong to a family of RSS referred to as symmetric spaces of noncompact type or noncompact RSS. 
In the following, we refer to noncompact RSS as RSS or symmetric spaces. 
Let $G$ be a connected noncompact semisimple Lie group with finite center, 
%and $\mathfrak{g}=\mathfrak{l}+\mathfrak{a}+\mathfrak{n}$ an Iwasawa decomposition of $\mathfrak{g}$. 
%Let K, A, and N denote the Lie subgroups corresponding to the Lie subalgebras $\mathfrak{l},\mathfrak{a}$, and $\mathfrak{n}$. 
and let $K$ be a maximal compact subgroup of $G$. Then a symmetric space $X$ consists of the left cosets
\begin{equation*}
X := G/K := \{ x = gK | g \in G \}.
\end{equation*}

The action of $G$ on $X = G/K$ is defined as
%\begin{equation*}
$g[x] = g[hK] = ghK$ for $x = hK \in X$, $g,h \in G$. 
%\end{equation*}
Let $o$ be the origin $K$ in $X$, %Since $K$ is the isotropy subgroup of $G$ at $o$, 
then the map $\gamma:gK \mapsto g[o]$ is a diffeomorphism of $G/K$ onto $X$. 

Let $d(.,.)$ be the distance induced by the Riemannian metric. 
A \textit{geodesic ray} in $X$ is a map $\delta: [0, \infty) \rightarrow X$ such that 
$d(\delta(t),\delta(t')) = |t - t'|,\forall t,t' \ge 0$. 
A \textit{geodesic line} in $X$ is a map $\delta: \mathbb{R} \rightarrow X$ 
such that $d(\delta(t), \delta(t')) = |t - t'|, \forall t,t' \in \mathbb{R}$. 

The geometry of $X$ can be studied through the geometry of its 
\textit{maximal flats}~\cite{ballmann2012,bridson2011metric,helgason1979differential,kapovich2017anosovsubgroupsdynamicalgeometric}. 
A subspace $F \subset X$ is called a flat of dimension $k$ (or a $k$-flat) if it is isometric to $\mathbb{R}^k$. 
The subspace $F$ is called a maximal flat if it is not contained in any flat of bigger dimension. 
Since all maximal flats in $X$ are isometric~\cite{helgason1979differential}, 
they can be simultaneously identified with a model (maximal) flat $F_{mod}$. 

Flats are decomposed into \textit{Weyl chambers}. 
A Weyl chamber in a maximal flat $F$ with tip at $x \in F$ 
is a connected component of the set of points $x' \in F \setminus \{x\}$ such that the geodesic line through $x$ and $x'$ 
is contained in a unique maximal flat~\cite{bridson2011metric}. 
Since $G$ acts transitively on the set of Weyl chambers in $X$~\cite{helgason1979differential}, 
they can be simultaneously identified with a Weyl chamber $\Delta$. 
The subgroup of isometries of $F$ which are induced by elements of $G$ is isomorphic to a semidirect product $\mathbb{R}^r \rtimes W$. 
$W$ is called the \textit{Weyl group} of $G$ and $X$. 

Any symmetric space $X$ is associated with a \textit{boundary at infinity} $\partial X$ constructed as the set of 
equivalence classes of geodesic rays in $X$. 
Two rays are considered equivalent if their images are a bounded distance apart~\cite{bridson2011metric}. 
The equivalence class of a geodesic ray $\delta$ is denoted by $\delta(\infty)$. 

\section{Proposed Approach}
\label{sec:proposed_approach}

Our proposed point-to-hyperplane distances based on the quotient structure of Siegel spaces and the vector-valued distance
are presented in Sections~\ref{sec:hyperplanes_quotient_structure} and~\ref{sec:hyperplanes_vvd}, respectively. 
In Section~\ref{sec:app_siegel_neural_networks}, we present our MLR models and introduce two variants of FC layers for 
Siegel neural networks. In our work, we focus on Siegel spaces but many of our results can also be stated for other RSS. 
To simplify the notation, we use the letters $X$, $G$, and $K$ (see Section~\ref{sec:symmetric_space_theory}) 
to denote the spaces associated with Siegel spaces (see Section~\ref{sec:siegel_spaces_symspaces}) unless otherwise stated. 

%-----------------------------------------------------------------------------------------------------------
\subsection{Point-to-hyperplane Distances Based on the Quotient Structure of Siegel Spaces}
\label{sec:hyperplanes_quotient_structure}

%-----------------------------------------------------------------------------------------------------------
\subsubsection{Hyperplanes}
\label{sec:hyperplane_quotient}

We start with a formulation of Euclidean hyperplanes given as 
%$\mathcal{H}^E_{a_j,b_j}$ is defined as
\begin{equation*}%\label{eq:euclidean_hyperplane_equation}
\mathcal{H}^E_{a,b} = \{ x \in \mathbb{R}^m: \langle x, a \rangle - b = 0 \},
\end{equation*}
where $a \in \mathbb{R}^m \setminus \{ \mathbf{0} \}$, $b \in \mathbb{R}$, 
and $\langle \cdot,\cdot \rangle$ is the Euclidean inner product. Hyperplane $\mathcal{H}^E_{a,b}$ 
can be reformulated~\cite{NEURIPS2018_dbab2adc} as
\begin{equation}\label{eq:reformulated_euclidean_hyperplane_equation}
\mathcal{H}^E_{a,b} = \{ x \in \mathbb{R}^m: \langle -p + x, a \rangle = 0 \},
\end{equation}
where $p \in \mathbb{R}^m$ and $\langle p, a \rangle = b$. 
 
To generalize Euclidean hyperplanes to our setting, 
we follow the approach in~\cite{NguyenGyroMatMans23,NguyenICLR24} which relies on a binary operation, 
an inverse operation, and an inner product defined on the target space. 
Let $x = gK, y = hK \in X$ where $g,h \in G$. In the case of $\mathbb{SH}_m$, $g=\phi(x), h=\phi(y)$ where $\phi(u+iv) = \begin{bmatrix} v^{\frac{1}{2}} & uv^{-\frac{1}{2}} \\ \mathbf{0} & v^{-\frac{1}{2}} \end{bmatrix}, u+iv \in \mathbb{SH}_m$. 

%\begin{definition}\label{def:loop_binary_opt}
\begin{definition}[\cite{NguyenICLR25}]\label{def:loop_binary_opt}
%%%For $x = gK, y = hK \in X$, $g,h \in G$, the binary operation $\oplus$ is defined as
The binary operation $\oplus$ and inverse operation $\ominus$ are defined as
\begin{equation*}
x \oplus y = ghK, \hspace{3mm} \ominus x = g^{-1}K. 
\end{equation*}
\end{definition}  

We propose the following inner product. 
\begin{definition}\label{def:inner_product_symspace}
%Let $x = gK, y = hK \in X$, $g,h \in G$.  
The inner product $\langle \cdot,\cdot \rangle_{\mathbb{S}}$ on $X$ is defined as
\begin{equation*}
\langle x,y \rangle_{\mathbb{S}} = \langle \operatorname{log}(gg^T),\operatorname{log}(hh^T) \rangle,
\end{equation*}
%where $\operatorname{Log}(.)$ is the Riemannian exponential map. 
where %$\langle \cdot,\cdot \rangle$ is the Euclidean inner product, %induced by the Killing form of $\mathfrak{g}$, 
$\operatorname{log}(\cdot)$ denotes the matrix logarithm. 
\end{definition}
 
Our proposed inner product is motivated by Proposition~\ref{lem:main_inner_product_prop}. 
(see Appendix~4.1 for its proof).  
\begin{proposition}\label{lem:main_inner_product_prop}
%Let $x = gK, y = hK \in X$, $g,h \in G$. %, and let $\langle .,. \rangle_{\mathbb{S}}$ be 
%the inner product as given in Definition~\ref{def:inner_product_symspace}. 
The inner product $\langle \cdot,\cdot \rangle_{\mathbb{S}}$ agrees with the Riemannian distance, i.e., 
\begin{equation*}
\| \ominus x \oplus y \|_{\mathbb{S}} \propto d_{\mathbb{SH}}(x,y),
\end{equation*} 
where $x,y \in X$, and the norm $\| \cdot \|_{\mathbb{S}}$ is induced by the inner product $\langle \cdot,\cdot \rangle_{\mathbb{S}}$. 
Furthermore, 
the inner product $\langle \cdot,\cdot \rangle_{\mathbb{S}}$ is invariant under the action of K, i.e., for any $k \in K$, 
\begin{equation*}
\langle x,y \rangle_{\mathbb{S}} = \langle k[x],k[y] \rangle_{\mathbb{S}}.
\end{equation*}
\end{proposition}

Note that both properties in Proposition~\ref{lem:main_inner_product_prop} are satisfied by the inner products 
in~\cite{NguyenGyroMatMans23,NguyenICLR25} and the second property is also satisfied by the one 
in~\cite{helgason1994geometric}. 
Note also that these properties hold for the more general case in which $G$ is 
the general linear group or its subgroup, and $K$ is the group of orthogonal matrices or its subgroup (see Appendix~4.1). 
We are now ready to define hyperplanes. 

%\begin{definition}[{\bf Hypergyroplanes}]\label{def:hypergyroplanes}
\begin{definition}\label{def:hypergyroplanes}
Let $a, p \in X$. Then hyperplanes on $X$ are defined as
\begin{equation*}%\label{eq:spd_hyperplanes}
\mathcal{H}_{a,p} = \{ x \in X: \langle \ominus p \oplus x, a \rangle_{\mathbb{S}} = 0 \}.         
\end{equation*}
%where $\langle .,. \rangle_{\mathbb{S}}$ denotes the inner product on $X$. 
\end{definition}

Segments of the form $\ominus p \oplus x$ can be regarded as Siegel analogs of Euclidean lines. 
Thus, $\mathcal{H}_{a,p}$ has a similar interpretation as a Euclidean hyperplane, i.e., 
the former contains a fixed point $p \in X$ and any point $x \in X$ such that 
the segment $\ominus p \oplus x$ is orthogonal to a fixed direction $a$.  
Therefore, hyperplanes as given in Definition~\ref{def:hypergyroplanes} are natural extensions of Euclidean hyperplanes.

%--------------------------------------------------------------------------------------------------------------------------------------
\subsubsection{Point-to-hyperplane Distance}
\label{sec:point_to_hyperplane_distance_quotient}

The distance $\bar{d}(x,\mathcal{H}_{a,p})$ between a point $x \in X$ and a hyperplane $\mathcal{H}_{a,p}$ 
given in Definition~\ref{def:hypergyroplanes} can be formulated~\cite{NguyenGyroMatMans23} as
\begin{equation*}
\bar{d}(x,\mathcal{H}_{a,p}) = \sin(\angle x p \bar{q}) d(x,p),
\end{equation*}
where $\angle x p \bar{q}$ is the gyroangle~\cite{NguyenGyroMatMans23,UngarHyperbolicNDim} (see Appendix~3.4) 
between $\ominus p \oplus x$ and $\ominus p \oplus \bar{q}$, 
and $\bar{q}$ is computed as
\begin{align*}
\begin{split}
\bar{q} = \argmax_{q \in \mathcal{H}_{a,p} \setminus \{ p \}} \bigg( \frac{\langle \ominus p \oplus q, \ominus p \oplus x \rangle_{\mathbb{S}}}{\| \ominus p \oplus q \|_{\mathbb{S}}  \| \ominus p \oplus x \|_{\mathbb{S}}} \bigg),
\end{split}
\end{align*}

By convention, $\sin(\angle x p q) = 0$ for any $x,q \in \mathcal{H}_{a,p}$. 
Theorem~\ref{theorem:distance_to_hyperplanes} gives a closed form for the point-to-hyperplane distance on Siegel spaces 
(see Appendix~4.2 for its proof). 

\begin{theorem}\label{theorem:distance_to_hyperplanes}
Let $x, a, p \in X$ and 
let $\mathcal{H}_{a,p}$ be a hyperplane as given in Definition~\ref{def:hypergyroplanes}. 
Then %the distance $\bar{d}(x,\mathcal{H}_{a,p})$ between point $x$ and hyperplane $\mathcal{H}_{a,p}$ is given by
\begin{equation*}
\bar{d}(x,\mathcal{H}_{a,p}) = \frac{| \langle \operatorname{log}(\phi(p)^{-1} \phi(x) \phi(x)^T \phi(p)^{-T}), \operatorname{log}(\phi(a) \phi(a)^T) \rangle |}{ \| \operatorname{log}(\phi(a) \phi(a)^T) \| },
\end{equation*} 
where $\|\cdot\|$ denotes the Euclidean norm, and the map $\phi(\cdot)$ is given in Section~\ref{sec:siegel_spaces_symspaces}.   
\end{theorem}

%-----------------------------------------------------------------------------------------------------------
\subsubsection{Product Spaces}
\label{sec:mlr_product_spaces}

We now extend the above method to the product space setting. 
Let $X$ be defined as the Cartesian product $X = X_1 \times \ldots \times X_L$, 
where $X_j = G_j/K_j,j=1,\ldots,L$ are RSS, $G_j$ is a connected noncompact semisimple Lie group with finite center, 
$K_j$ is a maximal compact subgroup of $G_j$. Here we focus on the Cartesian product of SPD and Siegel spaces. 
%Other cases are left for future work.  
Each point $x \in X$ can be described through its coordinates 
$x=(x_1,\ldots,x_L),x_j \in X_j,j=1,\ldots,L$. 
%The tangent space $T_x X$ of $X$ at $x$ is then given as $T_x X = T_{x_1} X_1 \times \ldots \times T_{x_L} X_L$. 
In this setting, one has simple decompositions of the tangent space, the exponential map, and the squared Riemannian distance~\cite{FickenRiemannianProductSpaces39,GuMixedCurvatureICML19,TuragaRiemannianCV15}. 
When $X_j$ is an SPD space, $G_j$ is the general linear group and $K_j$ is the group of orthogonal matrices (see Appendix~3.3). 
Thus one can define the binary operation, inverse operation, and inner product on $X_j$ as in 
Definitions~\ref{def:loop_binary_opt} and~\ref{def:inner_product_symspace}, 
and the results in Proposition~\ref{lem:main_inner_product_prop} still hold.  
By abuse of notation, we shall use the same notations for 
those operations %on $X_j,j=1,\ldots,L$ 
as in Section~\ref{sec:hyperplane_quotient}. 

\begin{definition}\label{def:loop_binary_opt_product_spaces}
Let $x=(x_1,\ldots,x_L)$, $y=(y_1,\ldots,y_L) \in X,x_j,y_j \in X_j,j=1,\ldots,L$. 
The binary operation $\oplus$ and inverse operation $\ominus$ on $X$ are defined as
\begin{equation*}
x \oplus y = (x_1 \oplus y_1,\ldots,x_L \oplus y_L), \hspace{1mm} \ominus x = (\ominus x_1,\ldots,\ominus x_L).
\end{equation*}
\end{definition}

\begin{definition}\label{def:inner_product_symspace_product_spaces}
The inner product $\langle \cdot,\cdot \rangle_{\mathbb{S}}$ on $X$ is defined as
\begin{equation*}
\langle x,y \rangle_{\mathbb{S}} = \sum_{j=1}^L \langle x_j,y_j \rangle_{\mathbb{S}}.
\end{equation*}
\end{definition}

The following theorem (see Appendix~4.3 for its proof) extends Theorem~\ref{theorem:distance_to_hyperplanes} to the considered setting. 
\begin{theorem}\label{theorem:distance_to_hyperplanes_product_manifolds}
Let $\mathcal{H}_{a,p}$ be a hyperplane as given in Definition~\ref{def:hypergyroplanes},  
where $a=(a_1,\ldots,a_L), p=(p_1,\ldots,p_L), a_j = w_jK_j, p_j = h_jK_j \in X_j,w_j,h_j \in G_j,j=1,\ldots, L$,   
and let $x= (x_1,\ldots,x_L) \in X$ where $x_j = g_jK_j \in X_j,g_j \in G_j$.  
Then 
\begin{equation*}%\label{eq:distance_set_to_hyperplane_ai}
\bar{d}(x,\mathcal{H}_{a,p}) = \frac{| \sum_{j=1}^L \langle \operatorname{log}(h_j^{-1} g_j g_j^T h_j^{-T}), \operatorname{log}(w_j w_j^T) \rangle |}{\sqrt{ \sum_{j=1}^L \| \operatorname{log}(w_j w_j^T) \|^2} }. 
\end{equation*} 
\end{theorem}

%---------------------------------------------------------------------------------------------------------------------------
\subsection{Point-to-hyperplane Distances Based on Vector-Valued Distances}
\label{sec:hyperplanes_vvd}

As shown in~\cite{shimizu2021hyperbolic}, the formulation of Euclidean hyperplanes in Eq.~(\ref{eq:reformulated_euclidean_hyperplane_equation}) 
has an over-parameterization issue, i.e., it increases the number of parameters from $m+1$ to $2m$ in each class. 
Our formulation of hyperplanes in Section~\ref{sec:hyperplane_quotient} (see Definition~\ref{def:hypergyroplanes}) 
follows that formulation and thus suffers from a similar issue. 
In this section, we propose another method for constructing the point-to-hyperplane distance 
which results in more compact MLR layers for Siegel neural networks. 

%---------------------------------------------------------------------------------------------------------------------------
\subsubsection{Hyperplanes}
\label{sec:hyperplane_vvd} 

We start with a similar formulation of Euclidean hyperplanes in Eq.~(\ref{eq:reformulated_euclidean_hyperplane_equation}) but 
use a different parameterization. Given $p \in \mathbb{R}^m$ and $\xi \in \partial \mathbb{R}^m$, 
the Euclidean hyperplane $\mathcal{H}^E_{\xi,p}$ parameterized by $p$ and $\xi$ 
can be defined~\cite{NguyenICLR25} by
\begin{equation*}%\label{eq:euclidean_hyperplane_vvd_formulation}
\mathcal{H}^E_{\xi,p} = \{ x \in \mathbb{R}^m: \langle p - x, a \rangle = 0 \} = \{ x \in \mathbb{R}^m: \langle \operatorname{vec}(x,p), a \rangle = 0 \},
\end{equation*}
where %$a \in \mathbb{R}^m \setminus \{ \mathbf{0} \}$ is such that %$p \in \mathbb{R}^m$ and %$\langle p, a \rangle = b$.
$\xi$ is the equivalence class of the geodesic ray $\delta(t)=t\frac{a}{\|a\|}, a \in \mathbb{R}^m \setminus \{ \mathbf{0} \}$, 
and the function $\operatorname{vec}(x,p)=p-x$ denotes the translation carrying $x$ to $p$. 
 
In a symmetric space, a natural analog of the function $\operatorname{vec}(\cdot,\cdot)$ is the vector-valued distance 
function~\cite{kapovich2005generalizedtriangleinequalitiessymmetric,kapovich2017anosovsubgroupsdynamicalgeometric}. 
Given two points $x,y \in X$, one computes a $G$-invariant distance by first transforming 
(via the $G$-action) $x$ and $y$ to $x'$ and $y'$ on the model flat $F_{mod}$, respectively, and 
then identifying the translation modulo the action of the Weyl group %\footnote{More precisely, the image of the translation by a group action.} 
carrying $x'$ to $y'$. 
Note that in $\mathbb{R}^m$, the projections of $x$ and $p$ on a maximal flat are precisely $x$ and $p$, respectively. 
The domain of the resulting distance function, which is a fundamental domain for the action of the Weyl group on the translations, 
can be canonically identified with the Weyl chamber $\Delta$.  
The above observation motivates the following definition. 

\begin{definition}\label{def:symspace_hyperplanes_vvd}
%Let $p \in X$ and $\xi \in \partial X$. 
%Let $\theta$ be the $\Delta_{sph}$-direction map on $X$. %(see Section~\ref{sec:symmetric_space_theory}). 
% \olsi{\mathfrak{a}^+}
%Let $\Delta$ be the (closed) Weyl chamber and 
Let $d_{\Delta}(\cdot,\cdot):X \times X \rightarrow \Delta$ be the vector-valued distance function on $X$. 
Let $p \in X$, $\xi \in \partial X$, and let $a_{\xi} \in \Delta$ be such that $\xi$ is the equivalent class of the geodesic ray 
$\delta(t) = k\exp(ta_{\xi})K,k \in K$. 
Then hyperplane $\mathcal{H}_{\xi,p}$ is defined as
\begin{equation*}
\mathcal{H}_{\xi,p} = \{ x \in X: \left\langle d_{\Delta}(x,p),a_{\xi} \right\rangle = 0 \}. 
\end{equation*}
%where $\ominus$ and $\oplus$ are the inverse and binary operations on $X$, respectively.  
\end{definition}

A hyperplane given in Definition~\ref{def:symspace_hyperplanes_vvd} has a clear interpretation, i.e., 
it contains a fixed point $p \in X$ and any point $x \in X$ such that the vector-valued distance between $x$ and $p$ 
is orthogonal to a fixed direction $a_{\xi}$. 
We note that the notion of vector-valued distance has been employed in~\cite{FedericoSymspaces21,FedericoGyrocalculusSPD21} for 
learning and visualizing embeddings in natural language processing and graph tasks. However, none of those works 
reveals the analogies discussed above for defining Siegel hyperplanes.  

%-----------------------------------------------------------------------------------------------------------
\begin{figure}[t]
  \begin{center}
    \begin{tabular}{c}      
      \includegraphics[width=0.7\linewidth, trim = 180 305 100 120, clip=true]{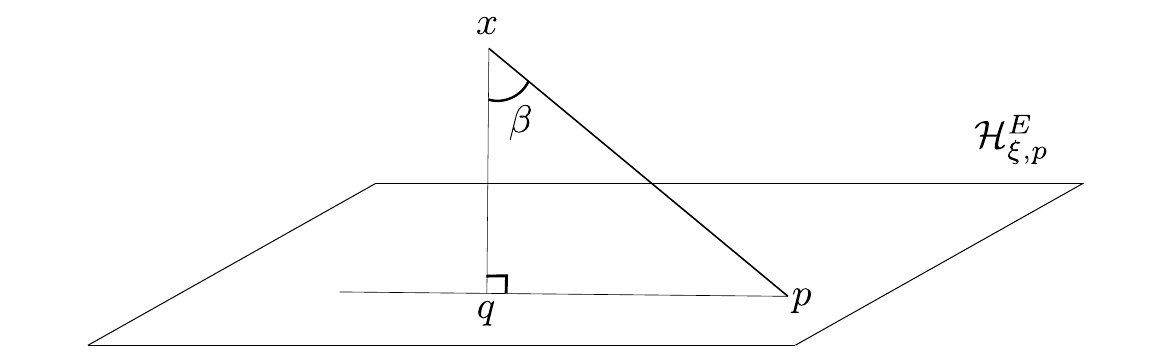} 
    \end{tabular}
  \end{center} 
  \caption{\label{fig:distance_to_hyperplane} The distance between a point $x \in \mathbb{R}^m$ and a Euclidean hyperplane $\mathcal{H}^E_{\xi,p}$. 
}    
\end{figure}
%-----------------------------------------------------------------------------------------------------------

%-----------------------------------------------------------------------------------------------------------
\subsubsection{Point-to-hyperplane Distance}
\label{sec:point_to_hyperplane_distance_vvd}
 
Let $p \in \mathbb{R}^m$ and $\xi \in \partial \mathbb{R}^m$. 
The distance $\bar{d}(x,\mathcal{H}^E_{\xi,p})$ between a point $x \in \mathbb{R}^m$ 
and hyperplane $\mathcal{H}^E_{\xi,p}$ can be computed (see Fig.~\ref{fig:distance_to_hyperplane}) as
\begin{equation*}%\label{eq:point_to_euclidean_hyperplane_distance}
\bar{d}(x,\mathcal{H}^E_{\xi,p}) = d(x,p)\cos(\beta),
\end{equation*}
where %$d(x,p)$ is the distance between $x$ and $p$, and 
$\beta$ is the angle between the segments $[x,p]$ and $[x,q]$, 
and $q$ is the projection of $x$ on $\mathcal{H}^E_{\xi,p}$. By convention, 
$\bar{d}(x,\mathcal{H}^E_{\xi,p}) = 0,\forall x \in \mathcal{H}^E_{\xi,p}$. 
%Let $\mathcal{H}^E_{\xi,p}$ be a hyperplane in $\mathbb{R}^m$. %passing through a point $p \in \mathbb{R}^m$ with a normal $\xi$. 
%Then Eq.~(\ref{eq:point_to_euclidean_hyperplane_distance}) can be rewritten as
The above equation can be rewritten as
\begin{equation*}
\bar{d}(x,\mathcal{H}^E_{\xi,p}) = d(x,p)\cos \angle_x(p,\xi),
\end{equation*}
where $\angle_x(p,\xi)$ denotes the angle at $x$ between the geodesic segment $[x, p]$ and 
the geodesic ray which issues from $x$ and is in the class $\xi$.  
We generalize the above equation to our setting. 
\begin{definition}\label{def:point_to_hyperplane_vvd}
Let $p \in X$, $\xi \in \partial X$, and let $\mathcal{H}_{\xi,p}$ be a hyperplane in $X$. %as given in Definition~\ref{def:symspace_hyperplanes_vvd}. 
%, and let $\| . \|_{\mathbb{S}}$ be a norm on $X$. 
Then the (signed) distance $\bar{d}(x,\mathcal{H}_{\xi,p})$ between a point $x \in X$ and hyperplane $\mathcal{H}_{\xi,p}$ is defined as
\begin{equation*}
\bar{d}(x,\mathcal{H}_{\xi,p}) = d(x,p)\cos \angle_x(p,\xi).
\end{equation*}
\end{definition}

Deriving a closed form of the point-to-hyperplane distance for applications from Definition~\ref{def:point_to_hyperplane_vvd} 
is not trivial. However, one can obtain an upper bound of 
this distance which is given in Proposition~\ref{prop:point_to_hyperplane_vvd_upper_bound} (see Appendix~4.4 for its proof). 

\begin{proposition}\label{prop:point_to_hyperplane_vvd_upper_bound}
% \olsi{\mathfrak{a}^+}
Let $x,p \in X$, $\xi \in \partial X$, and let $a_{\xi} \in \Delta$ be such that $\xi$ is the equivalent class of 
the geodesic ray $\delta(t) = k\exp(ta_{\xi})K$, where $k \in K$ and $\exp(\cdot)$ is the matrix exponential. 
%Let $\mathcal{H}_{\xi,p}$ be a hyperplane in $X$. %as given in Definition~\ref{def:symspace_hyperplanes_vvd}. 
%Let $x \in X$, and let $\bar{d}(x,\mathcal{H}_{\xi,p})$ be the distance between point $x$ and hyperplane $\mathcal{H}_{\xi,p}$.  
%as given in Definition~\ref{def:point_to_hyperplane_vvd}. 
Then 
\begin{equation*}
\bar{d}(x,\mathcal{H}_{\xi,p}) \le \left\langle d_{\Delta}(x,p),a_{\xi} \right\rangle. 
\end{equation*}
\end{proposition}

Note that the point-to-hyperplane distance in Section~\ref{sec:hyperplanes_quotient_structure} 
as well as those in~\cite{NEURIPS2018_dbab2adc,NguyenGyroMatMans23,NguyenICLR24} are obtained by solving 
an optimization problem in Euclidean spaces. This is different from our method in this section 
which estimates an upper bound of the point-to-hyperplane distance on the target spaces. 

%-----------------------------------------------------------------------------------------------------------------------------------
\subsection{Neural Networks on Siegel Spaces}
\label{sec:app_siegel_neural_networks}

In this section, we show how to construct MLR layers for Siegel neural networks using the tools introduced in 
Sections~\ref{sec:hyperplanes_quotient_structure} and~\ref{sec:hyperplanes_vvd}. 
We also propose two types of FC layers which are crucial building blocks in the context of deep neural networks. 

%-----------------------------------------------------------------------------------------------------------------------------------
\subsubsection{MLR Layers}
\label{sec:mlr_layers}

We follow the approach in~\cite{NEURIPS2018_dbab2adc,LebanonMarginClassifierICML04} for building Riemannian MLR. 
Given $M$ classes, (Euclidean) MLR computes the probability of each of the output classes as
\begin{align}\label{eq:mlr_reexpression}
\begin{split}
p(y=j|x) = \frac{\exp( a_j^Tx  - b_j)}{\sum_{j=1}^M \exp( a_j^Tx  - b_j)} & \propto \exp( a_j^Tx - b_j), % \\ &= \exp(\operatorname{sign}(\langle a_j,x \rangle - b_j) \| a_j \| d(x,\mathcal{H}_{a_j,b_j})),
\end{split}
\end{align}
where $x$ is an input sample,  
$b_j \in \mathbb{R}$, $x,a_j \in \mathbb{R}^m, j=1,\ldots,M$, %, and $\langle .,. \rangle$ denotes the dot product of two vectors. 
and $\exp(\cdot)$ is the ordinary exponential function (by abuse of notation). 
As shown in~\cite{LebanonMarginClassifierICML04}, Eq.~(\ref{eq:mlr_reexpression}) can be rewritten as
\begin{equation*}
p(y=j|x) \propto \exp(\operatorname{sign}( a_j^Tx - b_j) \| a_j \| \bar{d}(x,\mathcal{H}^E_{a_j,b_j})),  
\end{equation*}
where $\bar{d}(x,\mathcal{H}^E_{a_j,b_j})$ is the distance from point $x$ to hyperplane $\mathcal{H}^E_{a_j,b_j}$ 
(see Section~\ref{sec:hyperplane_quotient}). In our case, 
a hyperplane can be parameterized by two elements in $X$ (see Definition~\ref{def:hypergyroplanes}), 
or by an element in $X$ and an element in $\Delta$ (see Definition~\ref{def:symspace_hyperplanes_vvd}). 
We replace the expression in the argument of the function 
$\exp(\cdot)$ by the distances in Theorems~\ref{theorem:distance_to_hyperplanes} and~\ref{theorem:distance_to_hyperplanes_product_manifolds} 
as well as the upper bound of the point-to-hyperplane distance in Proposition~\ref{prop:point_to_hyperplane_vvd_upper_bound}. The final formulations of our MLR layers are given in Appendix~1. 

%-----------------------------------------------------------------------------------------------------------------------------------
\subsubsection{FC Layers}
\label{sec:fc_layers}

\paragraph{The FC layer with group action (AFC)}
Let $a+ib \in \mathbb{SH}_m$. 
%The group action mapping $iI_m$ to $a+ib$ is given by
Then the element in $\operatorname{Sp}_{2m}$ mapping $iI_m$ to $a+ib$ via the group action (see Section~\ref{sec:siegel_spaces_symspaces}) is given by
\begin{equation*}
\phi(a+ib) = \begin{bmatrix} b^{\frac{1}{2}} & ab^{-\frac{1}{2}} \\ \mathbf{0} & b^{-\frac{1}{2}} \end{bmatrix}.
\end{equation*}

Given an input $x \in \mathbb{SH}_m$, the output of the AFC layer is obtained by taking the group action 
$\phi(a+ib)[x]$. This leads us to the following construction. 
\begin{definition}\label{def:fc_layers_group_action}
Let $x = u + iv \in \mathbb{SH}_m$ be the input of the AFC layer. Then the output of the AFC layer is given by:
\begin{equation*}
t = (b^{\frac{1}{2}}ub^{\frac{1}{2}} + a) + ib^{\frac{1}{2}}vb^{\frac{1}{2}},
\end{equation*}
where $a \in \operatorname{Sym}_m$ and $b \in \operatorname{Sym}^+_m$ are the parameters of the layer. 
\end{definition}

We have that $b^{\frac{1}{2}}ub^{\frac{1}{2}} + a \in \operatorname{Sym}_m$ and 
$b^{\frac{1}{2}}vb^{\frac{1}{2}} \in \operatorname{Sym}_m^+$ by construction. 
Hence, the AFC layer always outputs points on $\mathbb{SH}_m$. The transformation performed by the AFC layer 
can be interpreted as a translation of the input $x$ by $a + ib$ (see Section~\ref{sec:hyperplane_quotient}). 

\paragraph{The FC layer for dimensionality reduction  (DFC)}
Based on the definition of the AFC layer, another type of FC layers for Siegel neural networks can also be built 
using a method similar to~\cite{HuangGool17}.
\begin{definition}\label{def:fc_layers_general}
Let $\operatorname{St}_{m,m_2}$ be the space of $m \times m_2$ real matrices ($m > m_2$) 
with mutually orthogonal columns of unit length (the compact Stiefel manifold), 
and let $x = u + iv \in \mathbb{SH}_m$ be the input of the DFC layer. Then the output of the DFC layer is given by:
\begin{equation*}
t = (b^Tub + a) + ib^Tvb,
\end{equation*}
where $a \in \operatorname{Sym}_{m_2}$ and $b \in \operatorname{St}_{m,m_2}$ are the parameters of the layer.
\end{definition}

Our FC layers generalize some FC layers in previous works. 
Specifically, when $u=0$ and $a=0$, the imaginary part $b^{\frac{1}{2}}vb^{\frac{1}{2}}$ of the output of the AFC layer 
corresponds to the transformation performed by the affine-invariant translation layer~\cite{NguyenGyroMatMans23}, 
and the imaginary part $b^Tvb$ of the output of the DFC layer corresponds to the transformation performed by the 
well-known Bimap layer~\cite{HuangGool17}. 
In~\cite{Sonoda2022FCRidgele}, the authors also proposed FC layers for neural networks on RSS. 
However, these layers are different from our FC layers in some aspects. First, the former include activation functions which are not used in the latter. Second, the former do not output points on the considered spaces, as opposed to the latter which always output points on these spaces. 

%-----------------------------------------------------------------------------------------------------------------------------------
\section{Related Work}
\label{sec:related_work}

Existing MLR models on Riemannian manifolds are generally built on either SPD manifolds~\cite{chen2024riemannian,NguyenGyroMatMans23} and their low-rank counterparts~\cite{NguyenICLR24} or 
hyperbolic spaces~\cite{BdeirFullyHnnCv24,NEURIPS2018_dbab2adc,LebanonMarginClassifierICML04,shimizu2021hyperbolic}. 
Many of them~\cite{NEURIPS2018_dbab2adc,NguyenGyroMatMans23,NguyenICLR24,shimizu2021hyperbolic} leverage the gyro-structures of the Poincar\'e ball and SPD manifolds. 
The work in~\cite{NguyenICLR25} proposes MLR and FC layers for neural network on RSS %which are applicable to Siegel spaces. 
which rely on the construction of Busemann functions. 
The work in~\cite{Sonoda2022FCRidgele} analyzes some existing hyperbolic and SPD neural networks from the perspective of harmonic analysis on RSS. It mainly concerns with a constructive proof of the universal approximation property of finite neural networks on RSS. 
Our method in Section~\ref{sec:hyperplanes_quotient_structure} is inspired by the works 
in~\cite{NEURIPS2018_dbab2adc,NguyenGyroMatMans23,NguyenICLR24} and focuses on Siegel spaces. 
Our method in Sections~\ref{sec:hyperplanes_vvd} explores the connection between the point-to-hyperplane distance and the vector-valued distance which has not been investigated in previous works. 

%-----------------------------------------------------------------------------------------------------------------------------------
\section{Experiments}
\label{sec:experiments}

This section reports results of our experiments on the radar clutter classification and node classification tasks. 
For further details, please refer to Appendix~1 in which we present more experimental results on human action recognition and Riemannian generative modeling. 

%------------------------------------------------------------------------------------------------------------------------------
\begin{table}[t]
\begin{center}
  \resizebox{0.96\linewidth}{!}{
  \def\arraystretch{1.2}
  \begin{tabular}{| l | c | c | c | c |}    
    \hline
    \multirow{2}{*}{Method} & Dataset 1 & Dataset 2 & Dataset 3 & Dataset 4 \\   
    & $(3,600,3600)$ & $(4,100,2000)$ & $(5,80,1600)$ & $(6,50,500)$ \\            
    \hline        
    kNN~\cite{CabanesClassifGSI21} & 76.22$\pm$0.0 & 93.00$\pm$0.0 & 76.75$\pm$0.0 & 73.20$\pm$0.0 \\ 
    SPDNet~\cite{HuangGool17} & 63.44$\pm$0.11 & 41.50$\pm$0.12 & 45.88$\pm$0.15 & 66.80$\pm$0.04 \\
    SPDNetBN~\cite{BrooksRieBatNorm19} & 62.67$\pm$0.10 & 45.10$\pm$0.08 & 45.75$\pm$0.15 & 68.40$\pm$0.04 \\    
    MLR-AI~\cite{NguyenGyroMatMans23} & 65.61$\pm$0.15 & 47.40$\pm$0.12 & 46.12$\pm$0.17 & 67.60$\pm$0.04 \\
    GyroSpd++~\cite{NguyenICLR24} & 62.24$\pm$0.16 & 46.20$\pm$0.14 & 48.25$\pm$0.19 & 67.80$\pm$0.08 \\    
    \hline    
    SiegelNet-DFC-QMLR$_{\operatorname{Sym}_m^+ \times \mathbb{SH}_m^{q-1}}$ (Ours) & 40.78$\pm$0.23 & 82.70$\pm$0.18 & 74.88$\pm$0.21 & 71.20$\pm$0.08 \\ 
    SiegelNet-AFC-QMLR$_{\operatorname{Sym}_m^+ \times \mathbb{SH}_m^{q-1}}$ (Ours) & {\bf 80.94$\pm$0.14} & {\bf 96.50$\pm$0.12} & {\bf 91.00$\pm$0.18} & {\bf 85.60$\pm$0.06} \\
    \hline	
  \end{tabular}
  } 
\end{center}
\caption{\label{tab:exp_time_series_classification} Results (mean accuracy $\pm$ standard deviation) computed over 10 runs for radar clutter classification. The tuple $(m,M,s)$ below each dataset indicates the signal dimension $m$, the number of classes $M$, and the size of the dataset $s$.}
\end{table}
%------------------------------------------------------------------------------------------------------------------------------

%-----------------------------------------------------------------------------------------------------------------------------------------
\subsection{Radar Clutter Classification}
\label{sec:radar_signal_classification}

Radar clutter classification aims at  
recognizing different types of radar clutter which is the information recorded by a radar related to 
seas, forests, fields, cities and other environmental elements surrounding the radar~\cite{CabanesThesis}. 
Due to the scarcity of publicly available radar datasets for the task, our experiments are performed using 
simulated radar signals\footnote{\url{https://github.com/nguyenxuanson10/synthetic-data}} 
which are commonly assumed to be stationary centered autoregressive (AR) Gaussian time series~\cite{BarbarescoSuperResolution96,BarbarescoInfoGeomCov2013,BillingsleyRadar,CabanesThesis}. The AR model is given by
\begin{equation*}
u_n + \sum_{j=1}^q c_j u_{n-j} = v_n,
\end{equation*}
where $q$ ($q > 1$) is the order of the AR model, 
$u_n \in \mathbb{C}^m$ is the vector of signals at time $n$, 
$c_j  \in \mathbb{C}^{m \times m},j=1,\ldots,q$ are the prediction coefficients (AR parameters), 
and $v_n \in \mathbb{C}^m$ is the prediction error at time $n$ which is assumed to be a multidimensional Gaussian random variable 
(detailed descriptions of the construction of our datasets are provided in Appendix~1.1). 
To compute an input data for our networks from a time series, 
we parameterize the time series as $(p_0,w_1,\ldots,w_{q-1}) \in \mathbb{H}_m^+ \times \mathbb{SD}_m^{q-1}$, 
where $p_0 \in \mathbb{H}_m^+$ and $w_1,\ldots,w_{q-1} \in \mathbb{SD}_m$ (see Appendix~1.1). 
We note that methods dealing with data that lie on these product spaces have already been studied in previous works~\cite{BarbarescoInfoGeomCov2013,CabanesThesis,CabanesClassifGSI21}. These representation spaces are endowed with a natural metric inspired by information geometry~\cite{BarbarescoInfoGeomCov2013,CabanesThesis}. 
We discard the imaginary part of the component $p_0$ and map it to an SPD matrix $\tilde{p}_0$ (see Appendix~1.1). 
Each component $w_i$ is converted to $z_i \in \mathbb{SH}_m$ using the inverse matrix Cayley transformation 
(see Section~\ref{sec:siegel_spaces_symspaces}).  
The input data is thus represented by point %the coordinates 
$(\tilde{p}_0,z_1,\ldots,z_{q-1}) \in \operatorname{Sym}_m^+ \times \mathbb{SH}_m^{q-1}$. %in the product space $\operatorname{Sym}_m^+ \times \mathbb{SH}_m^{q-1}$. 

Each of our networks consists of an FC (AFC or DFC) layer and a MLR layer built on the distance in Theorem~\ref{theorem:distance_to_hyperplanes_product_manifolds}. 
The sizes of the parameter $b$ in the DFC layer are set to $3 \times 2$, $4 \times 3$, $5 \times 3$, and $6 \times 4$ 
for the experiments on datasets 1, 2, 3, and 4, respectively. 
We compare our approach to the following methods: 
(1) k-Nearest Neighbors (kNN) based on the K\"{a}hler distance~\cite{CabanesClassifGSI21} which is among 
the very few works for supervised classification in the product space $\mathbb{H}_m^+ \times \mathbb{SD}_m^{q-1}$; 
and (2) state-of-the-art SPD neural networks~\cite{BrooksRieBatNorm19,HuangGool17,NguyenGyroMatMans23,NguyenICLR24} 
which use the real parts of the covariance matrices estimated from the time series as input data 
(the real parts are mapped to SPD matrices as above). 
We use default settings for SPD models as in the original papers (see Appendix~1.1).
Results in Tab.~\ref{tab:exp_time_series_classification} show that 
SiegelNet-AFC-QMLR$_{\operatorname{Sym}_m^+ \times \mathbb{SH}_m^{q-1}}$ yields the best performance in terms of mean accuracy across all the datasets. 
It is able to improve upon kNN, the second best method, by a margin of 4.71\%, 3.5\%, 14.25\%, and 12.39\% on 
datasets 1, 2, 3, and 4, respectively. 
There are large gaps in the performance of our models, yet in most cases, our worst model still outperforms SPD models by large margins. 
The results of our networks and kNN demonstrate the representation power of Siegel spaces 
in the considered application. This is also confirmed by our experiments (see Appendix~1.1) in which the performance 
of SiegelNet-AFC-QMLR$_{\operatorname{Sym}_m^+ \times \mathbb{SH}_m^{q-1}}$ drops drastically when the coordinates associated with the product 
space $\mathbb{SH}_m^{q-1}$ (i.e., $z_1,\ldots,z_{q-1}$) are removed from the input data. 

%------------------------------------------------------------------------------------------------------------------------------
\begin{table}[t]
\begin{center}
  \resizebox{0.7\linewidth}{!}{
  \def\arraystretch{1.2}
  \begin{tabular}{| l | c | c | c |}    
    \hline
    Method & Glass & Iris & Zoo \\          
    \hline  
    kNN~\cite{CabanesClassifGSI21} & 29.65$\pm$0.0 & 31.66$\pm$0.0 & 33.33$\pm$0.0 \\
    LogEig classifier~\cite{FedericoSymspaces21}  & 41.54$\pm$4.22 & 34.33$\pm$3.46 & 51.04$\pm$3.53 \\ 
    SiegelNet-BFC-BMLR~\cite{NguyenICLR25} & 41.12$\pm$3.86 & 37.26$\pm$2.53 & 48.12$\pm$3.08 \\
    \hline       
    SiegelNet-AFC-VMLR (Ours) & 42.06$\pm$4.23 & 36.94$\pm$3.68 & 50.86$\pm$3.26 \\ 
    SiegelNet-AFC-QMLR$_{\mathbb{SH}_m}$ (Ours) & {\bf 45.79$\pm$4.66} & {\bf 38.20$\pm$3.03} & {\bf 53.37$\pm$4.23} \\ 
    \hline	
  \end{tabular}
  } 
\end{center}
\caption{\label{tab:exp_node_classification} Results (mean accuracy $\pm$ standard deviation) computed over 10 runs for node classification.}
\end{table}
%------------------------------------------------------------------------------------------------------------------------------

%------------------------------------------------------------------------------------------------------------------------------
\begin{table}[t]
\begin{center}
  \resizebox{0.66\linewidth}{!}{
  \def\arraystretch{1.2}
  \begin{tabular}{| l | c | c | c |}    
    \hline
    Method & Glass & Iris & Zoo \\          
    \hline      
    SiegelNet-BMLR~\cite{NguyenICLR25} & 40.55$\pm$3.50 & 36.94$\pm$2.09 & 46.43$\pm$3.64 \\
    \hline    
    SiegelNet-VMLR (Ours) & 41.78$\pm$4.11 & 36.89$\pm$3.73 & 50.38$\pm$3.47 \\ 
    SiegelNet-QMLR$_{\mathbb{SH}_m}$ (Ours) & {\bf 42.61$\pm$3.26} & {\bf 37.52$\pm$2.54} & {\bf 52.00$\pm$4.78} \\    
    \hline	
  \end{tabular}
  } 
\end{center}
\caption{\label{tab:exp_node_classification_mlr} Comparison (mean accuracy $\pm$ standard deviation) of MLR models on Siegel spaces.}
\end{table}
%------------------------------------------------------------------------------------------------------------------------------

%-----------------------------------------------------------------------------------------------------------------------------------------
\subsection{Node Classification}
\label{sec:node_classification}

We perform node classification experiments on Glass, Iris, and Zoo datasets 
from the UCI Machine Learning Repository~\cite{Dua2019}\footnote{\url{https://archive.ics.uci.edu/datasets}}. 
Like~\cite{FedericoSymspaces21}, our main aim is to demonstrate the applicability of our approach on Siegel spaces, 
and we do not necessarily seek state-of-the-art results for the target task. 
 
To create input data which are graph node embeddings on Siegel spaces, we optimize a distance-based loss function~\cite{GuMixedCurvatureICML19,FedericoSymspaces21}. 
Given the distances $\{ d_G(j_1,j_2) \}_{j_1,j_2=1}^M$ between all pairs of connected nodes $j_1$ and $j_2$, the loss function is 
given by: 
\begin{equation*}
\mathcal{L}(x) = \sum_{j_1,j_2=1}^M \left | \bigg( \frac{d_{\mathbb{SH}}(x_{j_1},x_{j_2})}{d_G(j_1,j_2)} \bigg)^2 - 1 \right |,
\end{equation*}
where $x_{j_1}$ and $x_{j_2}$ are the node representations on the embedding space of nodes $j_1$ and $j_2$, respectively, 
and $d_{\mathbb{SH}}(\cdot,\cdot)$ is the distance function given in Section~\ref{sec:siegel_spaces_symspaces}. 
This loss function captures the average distortion. 
We use the cosine distance to compute 
a complete input distance graph from the original features of the data points~\cite{ChamiTreeNeurIPS20,FedericoSymspaces21}. 
After the node embeddings\footnote{\url{https://github.com/nguyenxuanson10/synthetic-data}} are learned, they are used as input features for all methods. 
In our experiments, the embedding dimension is set to $6$. 

Each of our networks consists of an AFC layer and a MLR (QMLR$_{\mathbb{SH}_m}$ or VMLR) layer. The QMLR$_{\mathbb{SH}_m}$ and VMLR layers are built using 
the distances in Theorem~\ref{theorem:distance_to_hyperplanes} and the upper bound of $\bar{d}(x, \mathcal{H}_{\xi,p})$ in Proposition~\ref{prop:point_to_hyperplane_vvd_upper_bound}, respectively. 
We compare our networks to the following methods: 
(1) kNN based on the distance function $d_{\mathbb{SH}}(\cdot,\cdot)$; 
(2) LogEig classifier~\cite{FedericoSymspaces21}; 
and (3) SiegelNet-BFC-BMLR which consists of an FC (BFC) layer and a MLR (BMLR) layer based on Busemann functions~\cite{NguyenICLR25}. 
Results in Tab.~\ref{tab:exp_node_classification} show that SiegelNet-AFC-QMLR$_{\mathbb{SH}_m}$ gives the best mean accuracies across all the datasets. 
In terms of mean accuracy, SiegelNet-AFC-VMLR surpasses SiegelNet-BFC-BMLR on Glass and Zoo datasets. 
SiegelNet-AFC-VMLR also surpasses the LogEig classifier on Glass and Iris datasets. 
Tab.~\ref{tab:exp_node_classification_mlr} reports the results of SiegelNet-BFC-BMLR and our networks without FC layers. 
It can be observed that our MLR models achieve higher (mean) accuracies than the BMLR model. 
Specifically, SiegelNet-QMLR$_{\mathbb{SH}_m}$ improves upon SiegelNet-BMLR by a margin of 2.06\%, 0.58\%, and 5.57\% on Glass, Iris, and Zoo datasets, respectively. 
SiegelNet-AFC-QMLR$_{\mathbb{SH}_m}$ is able to improve by 3.17\%, 0.67\%, and 1.36\% w.r.t. SiegelNet-QMLR$_{\mathbb{SH}_m}$ on 
Glass, Iris, and Zoo datasets, respectively, demonstrating the effectiveness of the AFC layer. 
Although SiegelNet-VMLR is outperformed by SiegelNet-QMLR$_{\mathbb{SH}_m}$, it is important to note that  
the model size of the former is about two times smaller than that of the latter (see Appendix~1.2). 

%------------------------------------------------------------------------------------------------------------------------
\section{Limitation of Our Approach}
\label{sec:limitation_our_approach}

A limitation of our method in Section~\ref{sec:hyperplanes_quotient_structure} is that our formulation of Siegel hyperplanes 
suffers from an over-parameterization issue. We alleviate this problem by reparameterizing Siegel hyperplanes 
as proposed in Section~\ref{sec:hyperplanes_vvd}. However, this new parameterization does not yield competitive performance compared to the original one.  

Our methods rely on operations on Siegel spaces which are generally expensive. Our method in Section~\ref{sec:radar_signal_classification} suffers from high computational cost in the setting of high-dimensional radar signals. Similarly, the loss function in Section~\ref{sec:node_classification} is based on the average distortion, for which the distances over all pairs of points must be computed during training. Since the computation of the Riemannian distance between two points on a Siegel space (see Section~\ref{sec:siegel_spaces_symspaces}) is based on eigenvalue decomposition, our method in Section~\ref{sec:node_classification} is computationally expensive when it comes to learning on large graphs. 

Like hyperbolic and SPD spaces, Siegel spaces are spaces of non-positive curvature. Therefore, our method in Section~\ref{sec:node_classification} does not allow isometric embeddings of graphs with a different curvature property, e.g., non-negative curvature. Although it can still be applied in this case, the learned node embedding may not preserve the curvature property of the embedded graph, leading to poor performance. Furthermore, like other graph embedding approaches, low-dimensional embeddings on Siegel spaces are not able to capture complex relationships within data which can affect the performance of our method. 

%------------------------------------------------------------------------------------------------------------------------
\section{Conclusion}

We have proposed Riemannian MLR and FC layers which enable the construction of effective Siegel neural networks.  
Our MLR layers are built upon the quotient structure of Siegel spaces and the concept of vector-valued distance on RSS. 
Our FC layers are based on the action of the real symplectic group on Siegel spaces. 
We have provided experimental evaluations demonstrating state-of-the-art performance of our approach in the radar clutter classification 
and node classification tasks. 

There are several potential improvements and extensions to Siegel neural networks that could be addressed as future work. 
Based on our experimental results, it can be observed that the DFC layer gives inferior performance compared to the AFC layer.  
It is therefore desirable to develop alternative layers for the DFC layer which are able to achieve better performance. 
Also, important building blocks such as convolutional layers, batch normalization layers, pooling layers, and attention layers 
are not studied in our work. Those are crucial to the development of effective deep Siegel neural networks. 


\section*{Acknowledgments}

We are grateful for the constructive comments and feedback from the anonymous reviewers. 

{\small
\bibliographystyle{ieee}
\bibliography{references} 
}

%%%%%%%%%%%%%%%%%%%%%%%%%%%%%%%%%%%%%%%%%%%%%%%%%%%%%%%%%%%%

\appendix

%\section{Technical Appendices and Supplementary Material}
%Technical appendices with additional results, figures, graphs and proofs may be submitted with the paper submission before the full submission deadline (see above), or as a separate PDF in the ZIP file below before the supplementary material deadline. There is no page limit for the technical appendices.

%%%%%%%%%%%%%%%%%%%%%%%%%%%%%%%%%%%%%%%%%%%%%%%%%%%%%%%%%%%%

\newpage
\section*{NeurIPS Paper Checklist}

\begin{enumerate}

\item {\bf Claims}
    \item[] Question: Do the main claims made in the abstract and introduction accurately reflect the paper's contributions and scope?
    \item[] Answer: \answerYes{} % Replace by \answerYes{}, \answerNo{}, or \answerNA{}.
    \item[] Justification: We clearly state the paper's scope in the abstract and introduction, and our contributions at the end of the introduction.

\item {\bf Limitations}
    \item[] Question: Does the paper discuss the limitations of the work performed by the authors?
    \item[] Answer: \answerYes{} % Replace by \answerYes{}, \answerNo{}, or \answerNA{}.
    \item[] Justification: Please refer to Section~6. 

\item {\bf Theory assumptions and proofs}
    \item[] Question: For each theoretical result, does the paper provide the full set of assumptions and a complete (and correct) proof?
    \item[] Answer: \answerYes{} % Replace by \answerYes{}, \answerNo{}, or \answerNA{}.
    \item[] Justification: We clearly state the assumptions for each theoretical result and provide all the proofs in Appendix.

    \item {\bf Experimental result reproducibility}
    \item[] Question: Does the paper fully disclose all the information needed to reproduce the main experimental results of the paper to the extent that it affects the main claims and/or conclusions of the paper (regardless of whether the code and data are provided or not)?
    \item[] Answer: \answerYes{} % Replace by \answerYes{}, \answerNo{}, or \answerNA{}.
    \item[] Justification: We provide all details needed to reproduce our experimental results in the main paper and Appendix.

\item {\bf Open access to data and code}
    \item[] Question: Does the paper provide open access to the data and code, with sufficient instructions to faithfully reproduce the main experimental results, as described in supplemental material?
    \item[] Answer: \answerNo{} % Replace by \answerYes{}, \answerNo{}, or \answerNA{}.
    \item[] Justification: The datasets used for our experiments will be made available upon acceptance of the paper. In the main paper and Appendix, we already give details on our experimental settings and implementation which would be sufficient to reproduce our experimental results. 

\item {\bf Experimental setting/details}
    \item[] Question: Does the paper specify all the training and test details (e.g., data splits, hyperparameters, how they were chosen, type of optimizer, etc.) necessary to understand the results?
    \item[] Answer: \answerYes{} % Replace by \answerYes{}, \answerNo{}, or \answerNA{}.
    \item[] Justification: We provide all details regarding our experiments in the main paper and Appendix. Those details would be sufficient for the reader to understand and reproduce our experimental results. 

\item {\bf Experiment statistical significance}
    \item[] Question: Does the paper report error bars suitably and correctly defined or other appropriate information about the statistical significance of the experiments?
    \item[] Answer: \answerNo{} % Replace by \answerYes{}, \answerNo{}, or \answerNA{}.
    \item[] Justification: We report mean accuracy and standard deviation over several runs for all competing methods.

\item {\bf Experiments compute resources}
    \item[] Question: For each experiment, does the paper provide sufficient information on the computer resources (type of compute workers, memory, time of execution) needed to reproduce the experiments?
    \item[] Answer: \answerYes{} % Replace by \answerYes{}, \answerNo{}, or \answerNA{}.
    \item[] Justification: We provide the information about the computer resources used for our experiments. We also provide a complexity analysis (memory and time) in Appendix~1.  
    
\item {\bf Code of ethics}
    \item[] Question: Does the research conducted in the paper conform, in every respect, with the NeurIPS Code of Ethics \url{https://neurips.cc/public/EthicsGuidelines}?
    \item[] Answer: \answerYes{} % Replace by \answerYes{}, \answerNo{}, or \answerNA{}.
    \item[] Justification: We already checked the NeurIPS Code of Ethics and think that our research conducted in the paper conform with it. 

\item {\bf Broader impacts}
    \item[] Question: Does the paper discuss both potential positive societal impacts and negative societal impacts of the work performed?
    \item[] Answer: \answerYes{} % Replace by \answerYes{}, \answerNo{}, or \answerNA{}.
    \item[] Justification: Please refer to Appendix~2. 
    
\item {\bf Safeguards}
    \item[] Question: Does the paper describe safeguards that have been put in place for responsible release of data or models that have a high risk for misuse (e.g., pretrained language models, image generators, or scraped datasets)?
    \item[] Answer: \answerNA{} % Replace by \answerYes{}, \answerNo{}, or \answerNA{}.
    \item[] Justification: We think this question is not applied to our paper. 

\item {\bf Licenses for existing assets}
    \item[] Question: Are the creators or original owners of assets (e.g., code, data, models), used in the paper, properly credited and are the license and terms of use explicitly mentioned and properly respected?
    \item[] Answer: \answerYes{} % Replace by \answerYes{}, \answerNo{}, or \answerNA{}.
    \item[] Justification: We properly cite all works and provide all links to codes and data used for our paper.  

\item {\bf New assets}
    \item[] Question: Are new assets introduced in the paper well documented and is the documentation provided alongside the assets?
    \item[] Answer: \answerNA{} % Replace by \answerYes{}, \answerNo{}, or \answerNA{}.
    \item[] Justification: We do not provide any assets in our work. 

\item {\bf Crowdsourcing and research with human subjects}
    \item[] Question: For crowdsourcing experiments and research with human subjects, does the paper include the full text of instructions given to participants and screenshots, if applicable, as well as details about compensation (if any)? 
    \item[] Answer: \answerNA{} % Replace by \answerYes{}, \answerNo{}, or \answerNA{}.
    \item[] Justification: Our paper does not deal with human subjects. 

\item {\bf Institutional review board (IRB) approvals or equivalent for research with human subjects}
    \item[] Question: Does the paper describe potential risks incurred by study participants, whether such risks were disclosed to the subjects, and whether Institutional Review Board (IRB) approvals (or an equivalent approval/review based on the requirements of your country or institution) were obtained?
    \item[] Answer: \answerNA{} % Replace by \answerYes{}, \answerNo{}, or \answerNA{}.
    \item[] Justification: Our paper does not deal with human subjects.

\item {\bf Declaration of LLM usage}
    \item[] Question: Does the paper describe the usage of LLMs if it is an important, original, or non-standard component of the core methods in this research? Note that if the LLM is used only for writing, editing, or formatting purposes and does not impact the core methodology, scientific rigorousness, or originality of the research, declaration is not required.
    %this research? 
    \item[] Answer: \answerNA{} % Replace by \answerYes{}, \answerNo{}, or \answerNA{}.
    \item[] Justification: We did not use LLMs in any process of our work (writing, coding, etc,)

\end{enumerate}

\end{document}